\documentclass[10pt,conference]{IEEEtran}
\IEEEoverridecommandlockouts
\usepackage{cite}
\usepackage{amsmath,amssymb,amsfonts}
\usepackage{algorithmic}
\usepackage{graphicx}
\usepackage{textcomp}
\usepackage{xcolor}
\usepackage{tikz}
\usepackage{hyperref}
\def\BibTeX{{\rm B\kern-.05em{\sc i\kern-.025em b}\kern-.08em
    T\kern-.1667em\lower.7ex\hbox{E}\kern-.125emX}}
\makeatletter
\def\ps@IEEEtitlepagestyle{%
	\def\@oddfoot{\mycopyrightnotice}%
	\def\@evenfoot{}%
}
\def\mycopyrightnotice{%
	{\footnotesize Copyright Notice 979-8-3503-0500-5/23/ \$31.00 ©2023 IEEE \hfill}
	\gdef\mycopyrightnotice{}
}
\begin{document}

\title{Addressing the Impact of Localized Training Data in Graph Neural Networks\\
	{
	}
}

\author{\IEEEauthorblockN{Akansha A}
	\IEEEauthorblockA{\textit{Department of Mathematics}\\
		\textit{Manipal Institute of Technology}\\
		Manipal Academy of Higher Education - 576104, India.\\
		akansha.agrawal@manipal.edu.}
}

\maketitle

\begin{abstract}
In the realm of Graph Neural Networks (GNNs), which excel at capturing intricate dependencies in graph-structured data, we address a significant limitation. Most state-of-the-art GNNs assume an in-distribution setting, limiting their performance on real-world dynamic graphs.  This article seeks to assess the influence of training GNNs using localized subsets of graphs. Such constrained training data could result in a model performing well in the specific trained region, but struggling to generalize and provide accurate predictions for the entire graph. Within the realm of graph-based semi-supervised learning (SSL), resource constraints frequently give rise to situations where a substantial dataset is available, yet only a fraction of it can be labeled. Such circumstances directly impact the model's efficacy. This limitation affects tasks like anomaly detection or spam detection when labeling processes are biased or influenced by human subjectivity. To tackle the challenges posed by localized training data, we approach the problem as an out-of-distribution (OOD) data issue by by aligning the distributions between the training data, which represents a small portion of labeled data, and the graph inference process that involves making predictions for the entire graph. We propose a regularization method to minimize distributional discrepancies between localized training data and graph inference, improving model performance on OOD data. Extensive tests on popular GNN models show significant performance improvement on three citation GNN benchmark datasets. The regularization approach effectively enhances model adaptation and generalization, overcoming challenges posed by OOD data\footnote{Codes are available at \url{https://github.com/Akanshaaga/Reg_APPNP}.}.

\end{abstract}


\section{Introduction}
Graph Neural Networks (GNNs) are versatile tools for analyzing graph-structured data, with applications in social network analysis \cite{dav_ajm-22a,fan-19a}, drug discovery \cite{gau_day_jam-21a,wie_koh-20a}, recommendation systems \cite{gao_wan-22a,chu_yao-22a,wan_yuy-22a}, and knowledge graph reasoning \cite{ye_kum_sin-22a,yas_ren_bos-21a,zha_yao-22a}. They handle tasks such as graph generation, community detection, anomaly detection, node classification \cite{xia_wan_dai-22a,oon_suz-19a}, graph classification, and link prediction, constituting some of their most prevalent and extensively studied applications.


Despite the notable achievements of GNNs in tasks like link prediction, node classification, and graph generation, it's crucial to recognize that the majority of cutting-edge GNNs are constructed based on the presumption of an in-distribution setup \cite{kip_wel-2016a,vel_pet_gui-2017a,ham_wil_jur-2017a,wij_wan-19a}. Within this context, it is assumed that both training and testing data stem from identical distributions. Nevertheless, practical implementation of this assumption becomes intricate when dealing with actual real-world graphs, such as population graphs and citation networks, or social networks. These graphs are often characterized by dynamic and evolving structures, diverse node attributes, and complex interactions. As a result, the performance of traditional GNN models tends to degrade when confronted with out-of-distribution (OOD) data. 

Effectively handling OOD data is vital for ensuring the practical usability of GNNs in real-world scenarios with high-stakes applications. These applications include  financial analysis \cite{yan_zha_zho-21a}, molecule prediction \cite{hu_fey_zit-20a}, autonomous driving \cite{lia_yan_hu-20a}, criminal justice \cite{aga_lak_zit-21a}, and pandemic prediction \cite{pan_nik_vaz-21a}. Various endeavors have been undertaken to tackle the out-of-distribution (OOD) issue in GNNs. For instance, approaches like data or graph augmentation \cite{yu_lia-23a, sui_wan_jia-22a, fen_wen_jie-20a} have been explored. These methods encompass introducing alterations to the node attributes or graph's structure during the training process, aiming to emulate scenarios representative of OOD conditions. By training GNN models on augmented data that covers a broader range of graph variations, the models can learn to be more robust and generalize better to OOD data. Disentanglement-Based Graph Models \cite{li_hao_ziw-22a, fan_wan_mo-2022a, li_hao_ziw-21a, zho_kut_rib-22a}: this approach focuses on designing novel GNN architectures that explicitly disentangle the propagation step from the non-linear transformations. These models aim to separate the node attributes and the graph structure to better capture and model the complex relationships within the graph. Learning Strategies \cite{li_hao_ziw-22b, buf_dav_pie-22a, wu_boj_ale-22a, sad_ma_li-21a, fen_he_tan-19a, wan_li_jin-22a, liu_hu_wan-22a, liu_jin_pan-22a}: various learning strategies are proposed to improve GNN performance on OOD data. These strategies involve adapting the training process or loss functions to account for the challenges posed by OOD scenarios. 


In this article, our objective is to assess the influence of training the model solely on a confined subset of the graph. This restricted training data may result in a model that performs well on the specific region it was trained on but fails to generalize and make accurate predictions for the entire graph. The inability of a GNN model trained on localized data to generalize effectively raises concerns in various domains. For instance, anomaly detection or spam detection \cite{liu_che_yan-18a, wan_lin_cui-19a} tasks can suffer greatly when the labeling process is biased or influenced by human subjectivity. In such scenarios, the model's performance may be inflated within the labeled region, but its predictions outside that region may be erroneous. This limitation undermines the reliability and practical applicability of the GNN model. To tackle the challenges posed by localized training data, we approach the problem as an OOD data issue. We recognize the need to align the distributions between the training data, which represents a small portion of labeled data, and the graph inference process that involves making predictions for the entire graph. Through the minimization of distributional discrepancy between these two constituents, our goal is to bolster the model's capacity for proficient generalization and precise extrapolation beyond the confines of the training subset.

To narrow the divide between localized training data and holistic graph inference, we introduce a regularization technique. Our regularization approach focuses on minimizing the distributional discrepancy between the two distributions, ensuring that the model can accurately predict beyond the localized training subset. By simulating a shift in the training data, we introduce bias during the labeling process as done in \cite{zhu-21a}, creating a more significant difference between the two distributions. Subsequently, the regularization is executed by minimizing the distributional disparity between the complete graph distribution and the training data distribution acquired through the introduced bias. This regularization technique guides the model to adapt and generalize effectively, mitigating the limitations posed by localized training data.


\textbf{Key observation.} Our observations reveal that the degradation in accuracy in traditional state-of-the-art GNNs \cite{kip_wel-2016a,vel_pet_gui-2017a,ham_wil_jur-2017a} can be attributed to the propagation of mismatched probability distributions. This issue becomes exacerbated with each layer, resulting in poor accuracy. Consequently, in the context of out-of-distribution (OOD) graph data, improving accuracy necessitates minimizing the discrepancy between domain-specific features before propagation in the classification task. In light of this, we find that models \cite{gas_etal-19a,liu_gao_ji-20a, fen_wen_jie-20a} which treat propagation and nonlinear transformation operations as separate steps tend to perform better on OOD graph data. 

\textbf{Our contributions.} Our contributions in this research are threefold. Firstly, the proposed technique demonstrates stability by consistently improving the average prediction accuracy while maintaining a small standard deviation. Secondly, the proposed regularization approach can be applied to any given Graph Neural Network (GNN) model, making it a versatile solution for addressing OOD data challenges across various domains. Lastly, in the experimental section, we conducted extensive tests on popular GNN models, applying our regularization strategy. The results showed a significant improvement in performance when handling OOD data. Specifically, we evaluated our approach on well-known citation networks and observed a significant increase in accuracy compared to the previously reported results in \cite{zhu-21a}. These findings highlight the efficacy of our regularization method in enhancing model performance and addressing the challenges posed by OOD data.

In the subsequent sections, we delve into a comprehensive exploration of the pertinent concepts and methodologies. In Section \ref{sec:2}, we provide an in-depth analysis of the existing literature, highlighting the various approaches proposed to tackle the challenges we address in this work. In Section \ref{sec:3}, we establish the necessary preliminaries to lay the foundation for our subsequent discussions. Section \ref{sec:4} presents a formal statement of the problem, elucidating its intricacies. Moving forward, in Section \ref{sec:5}, we introduce our novel framework, Reg-APPNP, designed to address the issues at hand. Sections \ref{sec:6} and \ref{sec:7} are dedicated to presenting our experimental setup, discussing the obtained results, and finally, drawing meaningful conclusions from our findings.

\section{Related Work} \label{sec:2}

Numerous methodologies have been suggested to tackle the challenge posed by OOD data within graph-based contexts. These approaches can be broadly classified into three distinct categories.

The initial category encompasses techniques related to data or graph augmentation. These methods introduce alterations to the node attributes or graph's structure throughout the training phase, aiming to replicate scenarios of being out-of-distribution. By training Graph Neural Network (GNN) models on augmented data that covers a broader range of graph variations, the models can learn to be more robust and generalize better to OOD data. Several studies have explored this approach and demonstrated its effectiveness \cite{yu_lia-23a, sui_wan_jia-22a, fen_wen_jie-20a}.

The second category focuses on disentanglement and causality-based GNNs. These approaches aim to design novel GNN architectures that explicitly separate the propagation step from the non-linear transformations. By disentangling the graph structure and node attributes, these models can better capture and model the complex relationships within the graph, leading to improved performance on OOD data \cite{li_hao_ziw-22a, fan_wan_mo-2022a, li_hao_ziw-21a, zho_kut_rib-22a}.

The third category comprises learning strategies specifically tailored for OOD scenarios. Various learning strategies have been proposed to enhance GNN performance on OOD data. These strategies include graph adversarial learning, graph self-supervised learning and graph invariant learning. Graph invariant learning aims to learn representations that are invariant to graph isomorphisms, while graph adversarial learning focuses on training GNNs that are robust against adversarial attacks. Graph self-supervised learning leverages unlabeled data to pretrain GNN models and learn meaningful representations. These learning strategies adapt the training process or loss functions to address the challenges posed by OOD scenarios \cite{li_hao_ziw-22b, buf_dav_pie-22a, wu_boj_ale-22a, sad_ma_li-21a, fen_he_tan-19a, wan_li_jin-22a, liu_hu_wan-22a, liu_jin_pan-22a}.


\section{Preliminaries}\label{sec:3}
\textbf{Notation.} We represent the graph as $G = (V, E)$, with $|V| = n$ indicating the node count and $|E| = m$ representing the edge count within the graph. Precisely, $V$ denotes the set of nodes, and $E$ signifies the set of edges comprising the graph. To signify the $1$-hop neighborhood of a node $u$ within the graph, we employ $N_u \subset V$, representing the nodes directly linked to $u$ by a single edge. Within this graph context, we utilize the adjacency matrix $A \in \{0,1\}^{n \times n}$, where each matrix entry corresponds to the presence or absence of an edge between nodes. A value of $1$ indicates an edge, while $0$ signifies its absence. This matrix captures the initial node features $x_u\in \mathbb{R}^d$, corresponding to attributes or traits associated with each node $u\in V$ in the graph. This framework allows for representation of various node properties. Furthermore, we employ the feature matrix $\textbf{X} \in \mathbb{R}^{n \times d}$, where $n$ signifies node count and $d$ represents feature vector dimensionality.

\textbf{Creating Biased Training Data.} In exploring the repercussions of localized training data on the node classification task, generating biased training data with controllable bias becomes essential. To achieve this, we leverage the personalized PageRank vector associated with each node, denoted as $\Pi = (I-(1-\alpha)\tilde{A})^{-1}$. In this equation, $\tilde{A} = D^{-\frac{1}{2}}(A+I)D^{-\frac{1}{2}}$ stands for the normalized adjacency matrix, where $D = \sum_{i=1}^{n}a_{ii}$ represents the degree matrix of $A+I = (a_{ij})$ \cite{gas_etal-19a}. Following the procedure outlined in \cite{zhu-21a}, we utilize this personalized PageRank vector to create training data with a predetermined bias.

\textbf{Central Moment Discrepancy (CMD).}  The CMD metric quantifies the direct disparity between two distributions, denoted as $p$ and $q$, by utilizing their central moments. This metric is formulated as follows:
$$d_{CMD}(p,q) = \dfrac{1}{|b-a|}\|E(p)-E(q)\|_2+\sum_{k=2}^{\infty}\|c_k(p)-c_k(q)\|_2,$$
where $a$ and $b$ represent the support of the joint distribution, $E(\cdot)$ is the expected values of the distributions and $c_k(\cdot)$ are their respective central moments, considering different orders $k$ of moments.

\textbf{Maximum Mean Discrepancy (MMD).} The MMD is a metric employed for quantifying distinctions between distribution means within a high-dimensional feature space established by a kernel function. It quantifies the dissimilarity between two probability distributions by comparing their expected values in the rich Hilbert space. The formula for MMD is:
$$d_{MMD}(p, q) = \|\mu_p - \mu_q\|_H,$$
where $p$ and $q$ are two probability distributions, $\mu_p$ and $\mu_q$ are their means in the Hilbert space $H$, and $\|\cdot\|_H$ represents the norm in that space.

\textbf{Graph Neural Network (GNN) Layer}:
A standard $l$-th layer in a Graph Neural Network can be defined as follows:
\begin{align*}
	h^{(l)}_u & = propagation\biggl\{x^{(l-1)}_u, \{x^{(l-1)}_v \mbox{ where } v \in N_u\}\biggr\}\\
	x^{(l)}_u &  = transformation\{(h^{(l-1)}_u)\},
\end{align*}
where $x_u^{(l)}$ signifies the feature vector of node $u$ at layer $l$, and $x_u^{(0)}$ stands for the initial feature representation of node $u$. The propagation step updates $x_u^{(l)}$ based on feature vectors of neighboring nodes in $N_u$, and the transformation step further processes the updated feature vector $h_u^{(l)}$. This sequence is iterated for a total of $K$ layers, yielding the final node feature matrix $\textbf{X}^{(K)}$ used for classification, where $\textbf{X}^{(0)} = \textbf{X}$.

\begin{tikzpicture}[node distance=1cm, auto]
		\node[draw, circle] (input) {$x^{(l-1)}_u$};
		\node[draw, rectangle, right of=input, xshift=2cm] (propagation) {$h^{(l)}_u$};
		\node[draw, circle, right of=propagation, xshift=2cm] (output) {$x^{(l)}_u$};
		
		\draw[->] (input) -- (propagation);
		\draw[->] (propagation) -- (output);
%
%
%
%
		
		\node[align=center] at (1.6, -1) {Propagation\\Function};
		\node[align=center] at (4.6, -1) {Transformation\\Function};
		
	\end{tikzpicture}

\section{Problem Formulation}\label{sec:4}
This study addresses the issue of distributional shift within the context of semi-supervised learning (SSL) on a single graph where only a limited portion is labeled. In traditional SSL, a prevailing strategy involves employing the cross-entropy loss function. This function quantifies the discrepancy between the predicted labels $\hat{y}_i$ generated by a graph neural network for each node $i$ and the true labels $y_i$. The cumulative loss $l$ is computed as the mean of individual losses across all training instances $M$:
\[ l = \frac{1}{M} \sum_{i=1}^{M} l(y_i, \hat{y}_i). \]
When both training and testing data originate from the same distributions, i.e., $P_{\text{train}}(X, Y) = P_{\text{test}}(X, Y)$, optimizing the cross-entropy loss on the training data ensures the classifier is appropriately calibrated to make precise predictions on the testing data. However, a significant challenge in machine learning emerges when there is a divergence between the distributions of the training and testing datasets, signified by $P_{\text{train}}(X, Y) \neq P_{\text{test}}(X, Y)$. To address this challenge, we focus on the distributional shift specifically within the output of the final hidden layer, referred to as $\hat{Y}$. Conventional learning theory postulates that the distribution of labels given the representations $\hat{Y}$ remains consistent for both training and test data, implying $P_{\text{train}}(Y|\hat{Y}) = P_{\text{test}}(Y|\hat{Y})$.

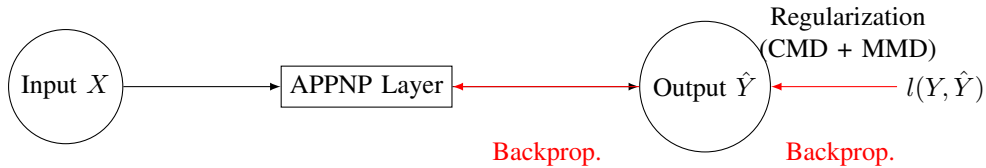
\begin{figure*}[ht]
	\centering
	\begin{tikzpicture}[node distance=2cm, auto]
		\node[draw, circle] (input) {Input $X$};
		\node[draw, rectangle, right of=input, xshift=2cm] (propagation) {APPNP Layer};
		\node[draw, circle, right of=propagation, xshift=2.5cm] (output) {Output $\hat{Y}$};
		\node (loss) [right of=output, xshift=1.2cm] {$l(Y,\hat{Y})$};
		
		\draw[-latex] (input) --  (propagation);
		\draw[-latex,black] (propagation) --node[midway, above] {} (output);
		\draw[-latex, red] (loss) --(output);
		\draw[-latex, red] (output) -- (propagation);
		\node[align=center] at (10.4, 0.7) {Regularization\\(CMD + MMD)};
		\node[align=center, red] at (10.3, -0.9) {Backprop.};
		\node[align=center, red] at (6.4, -0.9) {Backprop.};
	\end{tikzpicture}
	\caption{Illustration of Regularized APPNP (Reg-APPNP) Architecture. The regularization technique mitigates the impact of distributional shift in predicted labels. The model is trained end-to-end, ensuring comprehensive optimization.}
\end{figure*}



\section{Regularized Approximate personalized propagation of neural prediction (Reg-APPNP)}\label{sec:5}
In this segment, our focus lies on addressing the distributional shift challenge within graph neural networks. This scenario arises when the probability distributions of predicted labels deviate between the training and testing data, specifically denoted as $P_{\text{train}}(\hat{Y}) \neq P_{\text{test}}(\hat{Y})$. Leveraging our access to both the graph structure and unlabeled data, we introduce a regularization technique that harnesses two distinct discrepancy metrics: Central Moment Discrepancy (CMD) and Maximum Mean Discrepancy (MMD). These metrics enable us to quantitatively measure the dissimilarity between the probability distributions of the training and testing datasets.

\subsection{Reg-APPNP framework.}
We introduce our regularization method designed to minimize the discrepancy in the latent representations induced by distributional shift:
\begin{equation*}
	\begin{split}
		\mathcal{L}_{Reg-APPNP} = \dfrac{1}{M}\sum_{i=1}^{M}l(y_i,\hat{y}_i) + \lambda d_{CMD}(\hat{Y}_{train}, \hat{Y}_{test}) \\ +		
		\beta d_{MMD}(\hat{Y}{train}, \hat{Y}{test})
	\end{split}
\end{equation*}
Within this formulation, the regularization term encompasses the cross-entropy loss ($l$) that contrasts actual labels ($y_i$) with predicted labels ($\hat{y}_i$), alongside two discrepancy metrics: CMD and MMD. The parameters $\lambda$ and $\beta$ control the importance of each regularization component. Empirical results demonstrate that incorporating both CMD and MMD metrics as regularization techniques potentially enhances the regularization's effectiveness because they capture different aspects of distributional shift. CMD assesses the direct distance between distributions, focusing on their central moments that convey shape and variability information. On the other hand, MMD emphasizes differences in the means of distributions, capturing more global distinctions.
Our approach employs the APPNP model \cite{gas_etal-19a} on biased data with the proposed regularization term. Extensive experiments on real-world citation networks reveal that in some cases, one metric may dominate in capturing distributional shift, while in others, both metrics contribute equally. We fine-tune the parameters $\lambda$ and $\beta$ to achieve improved accuracy in the semi-supervised learning (SSL) classification task.
\section{Experiments.}\label{sec:6}
This section employs our introduced regularization approach on biased sample data, following the methodology outlined in Zhu et al.'s study \cite{zhu-21a}. We employ the identical validation and test divisions as established in the GCN paper \cite{kip_wel-2016a}. The objective is to assess the efficacy of our Reg-APPNP framework in mitigating distributional shifts as compared to alternative baseline methods. Furthermore, we present a comprehensive study on hyperparameter optimization specifically for the CORA dataset. Hyperparameters play a crucial role in the performance of our regularization technique, and we thoroughly investigate and optimize these parameters to achieve the best results.

\textbf{Datasets.} In our experimental analysis, we center our attention on the semi-supervised node classification task, employing three extensively utilized benchmark datasets: Cora, Citeseer, and Pubmed. For creating the biased training samples, we follow the data generation approach established for SR-GNN in Zhu et al.'s research \cite{zhu-21a}. To ensure equitable comparison, we adhere to the identical test splits and validation strategies employed in \cite{kip_wel-2016a}. In order to obtain unbiased data, we perform random sampling from the remaining nodes after excluding those assigned to the validation and test sets.

\textbf{Baselines.} To investigate the performance under distributional shift, we employ two types of GNN models. Firstly, we utilize traditional GNNs that incorporate message passing and transformation operations, including GCN \cite{kip_wel-2016a}, GAT \cite{vel_pet_gui-2017a}, and GraphSage \cite{ham_wil_jur-2017a}. Secondly, we explore another branch of GNNs that treats message passing and nonlinear transformation as separate operations. This includes models such as APPNP \cite{gas_etal-19a} and DAGNN \cite{liu_gao_ji-20a}.

\textbf{Our Method.} By default, we adopt the version of APPNP \cite{gas_etal-19a} augmented with our newly introduced regularization technique, which we term Reg-APPNP, as our foundational model. Furthermore, for validation purposes, we offer two ablations of Reg-APPNP. The first ablation solely incorporates the CMD metric as a form of regularization, while the second ablation solely integrates the MMD metric. Through these ablations, our objective is to gauge the efficacy of our proposed Reg-APPNP approach in mitigating distributional shifts and augmenting the performance of the graph neural network models.

\begin{figure}[ht]
	\centerline{\includegraphics[height=5.5cm,width=8cm]{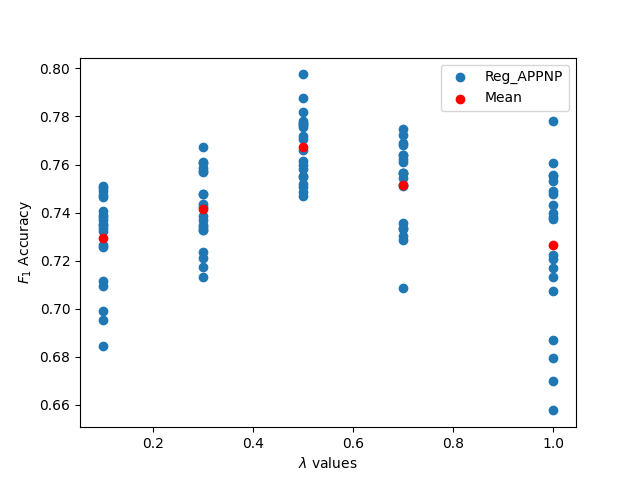}}
	\caption{Cumulative Impact of Penalty Parameter $\lambda$ (corresponding to $d_{CMD}$) on Accuracy.}
	\label{fig:lambda_parameter}
\end{figure}

\textbf{Hyperparameters.}The primary parameters in our methods are the penalty parameters $\lambda$ and $\beta$, which correspond to the regularization metrics $d_{CMD}$ and $d_{MMD}$, respectively. In light of empirical findings, we have identified the optimal parameter values for the SSL classification task across the three benchmark datasets. For the Cora dataset, we found that $\lambda = 0.5$ and $\beta = 1$ achieve the highest accuracy. In the case of Citeseer, the optimal values are $\lambda = 0.1$ and $\beta = 1$ , while for the Pubmed dataset, $\lambda = 0.1$ and $\beta = 0.1$ yield the best performance. Regarding the APPNP model, we set $\alpha = 0.1$, and we utilize the DGL (Deep Graph Library) layer for implementing the APPNP model within our experimental setup.  The outcomes, encompassing the influence of various $\lambda$ values on the Cora dataset, are illustrated in Figure \ref{fig:lambda_parameter}.
\subsection{Experimental Results} We begin by presenting a performance comparison between our proposed regularization technique and the shift-robust technique proposed by Zhu et al. \cite{zhu-21a}, when combined with popular GNN models trained on biased training data. Table \ref{table:reg_vs_sr} presents the F1-accuracy outcomes for semi-supervised node classification using the CORA dataset. The table demonstrates that our regularization technique consistently enhances the performance of each GNN model when applied to biased data. Notably, our method, Reg-APPNP, outperforms other baselines, including SR-GNN, on biased training data. 

This comparison confirms the efficacy of our proposed regularization technique in mitigating the impact of distributional shifts, leading to enhanced performance across various GNN models. Reg-APPNP emerges as the top-performing method, showcasing its effectiveness in handling biased training data and improving the accuracy of the semi-supervised node classification task.
\begin{table}[ht]
	\centering
	\caption{Performance comparison of our proposed regularization technique when combined with popular GNN models. The $F_1$-micro accuracy is reported and compared with the Shift-robust technique \cite{zhu-21a}.}
	\label{table:reg_vs_sr}
	\begin{tabular}{llll}
\hline \\[-0.7em]
Model & $F_1$-score & $F_1$-score w. Reg & $F_1$-score w. SR
\\\hline\\[-0.7em] \vspace{.1cm}
GCN & 65.35 $\pm$ $3.6$ & 71.41 $\pm$ $3.6$ & 70.45 $\pm$ $3.5$\\ \vspace{.1cm}
GraphSAGE & 67.56 $\pm$ $ 2.1$ & 72.10 $\pm $ $2.0$ & 71.79 $\pm $ $2.4$\\ \vspace{.1cm}
\textbf{APPNP} & 70.7 $\pm$ $ 1.9$  & \textbf{75.14 $\pm$ $1.8$} & 72.33 $\pm$ $ 5.9$\\ \vspace{.1cm}
DAGNN & 71.74 $\pm$ $ 0.32$  & 72.71 $\pm$ $ 0.31$ & 71.79 $\pm$ $ 0.36$	\\\vspace{.1cm}
\textbf{SR-GNN} & 73.5 $\pm $ $3.3$ \textbf{(\cite{zhu-21a})} &70.21 $\pm $ $0.87$	& 70.42 $\pm$ $ 1.1$ \\
\hline
	\end{tabular}
\end{table}

\begin{figure}[htbp]
	\centerline{\includegraphics[height=5.5cm,width=8cm]{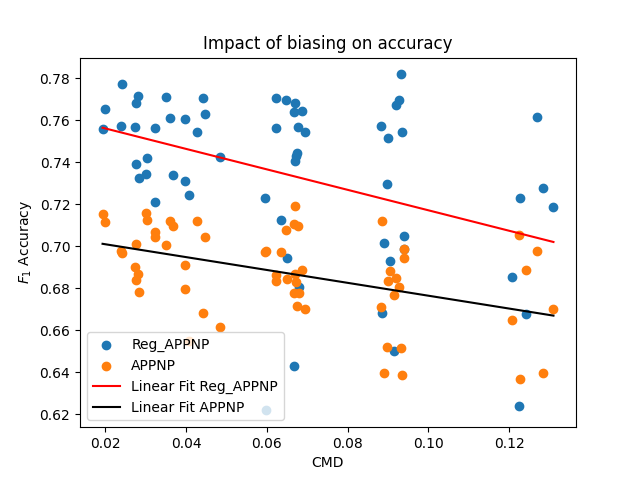}}
	\caption{Cumulative Impact of Biasing on Reg-APPNP and APPNP Models.}
	\label{fig:biasing_parameter}
\end{figure}

In the study conducted by the authors \cite{zhu-21a}, it was established that heightened bias levels lead to a degradation in the performance of prevalent GNN models. Illustrated in Figure \ref{fig:biasing_parameter}, we depict the behavior of our Reg-APPNP model as we escalate the bias within the training data. The outcomes distinctly reveal that our introduced regularization approach consistently enhances the performance in the semi-supervised node classification task, as opposed to the baseline APPNP model trained on biased data.

We employed our Reg-APPNP technique across three widely recognized citation GNN benchmarks and present the $F_1$-micro accuracy results in Table \ref{table:reg_appnp}. To evaluate the influence of biased training data, we contrast each model with the baseline APPNP trained on in-distribution IID data. Notably, as indicated in the second row of Table \ref{table:reg_appnp}, the performance of APPNP experiences a marked decline when trained on biased data.

However, our proposed Reg-APPNP demonstrates significant improvement in performance on biased data, outperforming SR-GNN \cite{zhu-21a} when we executed their code on our system. It's worth noting that the results for SR-GNN in the table differ from the reported ones for Citeseer and Pubmed. The values in the table correspond to the results we reproduced using their code on our personal system.

Finally, we observe that combining both metrics, $d_{CMD}$ and $d_{MMD}$, is more effective than using only one of them during the entropy loss step. This observation underscores the advantages of our holistic regularization approach in mitigating distributional shifts and enhancing the performance of graph neural networks under biased training data.
\begin{table}[ht]
	\centering
	\caption{Performance comparison of Reg-APPNP on citation network benchmarks. We report $F_1$-micro accuracy and compare with APPNP trained on in-distribution IID data.}
	\label{table:reg_appnp}
	\begin{tabular}{llll}
		\hline \\[-0.7em]
		Model & Cora & Citeseer & Pubmed \\\hline \\[-0.7em] \vspace{.1cm}
		APPNP (Unbiased) &85.09 $\pm$ $ 0.25$ &75.73 $\pm$ $ 0.30$ & 79.73 $\pm$ $ 0.31$\\\hline \\[-0.7em] \vspace{.1cm}
		APPNP & 70.7 $\pm$ $ 1.9$ & 60.78 $\pm$ $ 1.6$ & 53.42 $\pm$ $1.3$\\ \vspace{.1cm}
		Reg-APPNP w.o. CMD & 71.37 $\pm$ $ 1.2$ & 60.46 $\pm $ $1.9$ & 53.35 $\pm $ $1.0$\\ \vspace{.1cm}
		Reg-APPNP w.o. MMD & 75.03 $\pm$ $ 1.2$  & 62.72 $\pm$ $ 1.9$ &66.13 $\pm$ $ 1.2$ \\ \vspace{.1cm}
		Reg-APPNP(ours) & \textbf{75.14 $\pm$ $ 1.8$}  & \textbf{65.01 $\pm$ $ 2.0$} & \textbf{70.71 $\pm$ $ 2.9$}	\\\vspace{.1cm}
		{SR-GNN} & 73.5 $\pm $ $3.3$ &62.60 $\pm $ $0.6$	& 68.78 $\pm$ $ 2.2$ \\
		\hline
	\end{tabular}
\end{table}

\section{Conclusion}\label{sec:7} In our study, we noted that elevated bias levels lead to reduced performance in popular GNN models. We stressed the ubiquity of biased training data in real-world contexts, arising from factors like labeling complexities, varied heuristics, node selection inconsistencies, delayed labeling, and inherent problem constraints. To combat this issue, we introduce CMD and MMD as regularization terms within the APPNP model. This regularization incentivizes the model to minimize disparities in latent representations caused by distributional shifts. Our proposed regularization can be universally applied to any graph or GNN model, offering a versatile solution across diverse domains. This adaptability empowers researchers and practitioners to enhance model performance and generalization in the face of out-of-distribution scenarios.

\section*{Acknowledgment}
I extend my heartfelt appreciation to Dr. Karmvir Singh Phogat for providing invaluable insights and essential feedback on the research problem explored in this article. His thoughtful comments significantly enriched the quality and lucidity of this study.

\bibliographystyle{./IEEEtran}
\bibliography{./IEEEabrv,bibfile_GNN_original}

\begin{thebibliography}{10}
\providecommand{\url}[1]{#1}
\csname url@samestyle\endcsname
\providecommand{\newblock}{\relax}
\providecommand{\bibinfo}[2]{#2}
\providecommand{\BIBentrySTDinterwordspacing}{\spaceskip=0pt\relax}
\providecommand{\BIBentryALTinterwordstretchfactor}{4}
\providecommand{\BIBentryALTinterwordspacing}{\spaceskip=\fontdimen2\font plus
\BIBentryALTinterwordstretchfactor\fontdimen3\font minus
  \fontdimen4\font\relax}
\providecommand{\BIBforeignlanguage}[2]{{%
\expandafter\ifx\csname l@#1\endcsname\relax
\typeout{** WARNING: IEEEtran.bst: No hyphenation pattern has been}%
\typeout{** loaded for the language `#1'. Using the pattern for}%
\typeout{** the default language instead.}%
\else
\language=\csname l@#1\endcsname
\fi
#2}}
\providecommand{\BIBdecl}{\relax}
\BIBdecl

\bibitem{dav_ajm-22a}
A.~Davies and N.~Ajmeri, ``Realistic synthetic social networks with graph
  neural networks,'' \emph{arXiv preprint arXiv:2212.07843}, 2022.

\bibitem{fan-19a}
W.~Fan, Y.~Ma, Q.~Li, Y.~He, E.~Zhao, J.~Tang, and D.~Yin, ``Graph neural
  networks for social recommendation,'' in \emph{The world wide web
  conference}, 2019, pp. 417--426.

\bibitem{gau_day_jam-21a}
T.~Gaudelet, B.~Day, A.~R. Jamasb, J.~Soman, C.~Regep, G.~Liu, J.~B. Hayter,
  R.~Vickers, C.~Roberts, J.~Tang \emph{et~al.}, ``Utilizing graph machine
  learning within drug discovery and development,'' \emph{Briefings in
  bioinformatics}, vol.~22, no.~6, p. bbab159, 2021.

\bibitem{wie_koh-20a}
O.~Wieder, S.~Kohlbacher, M.~Kuenemann, A.~Garon, P.~Ducrot, T.~Seidel, and
  T.~Langer, ``A compact review of molecular property prediction with graph
  neural networks,'' \emph{Drug Discovery Today: Technologies}, vol.~37, pp.
  1--12, 2020.

\bibitem{gao_wan-22a}
C.~Gao, X.~Wang, X.~He, and Y.~Li, ``Graph neural networks for recommender
  system,'' in \emph{Proceedings of the Fifteenth ACM International Conference
  on Web Search and Data Mining}, 2022, pp. 1623--1625.

\bibitem{chu_yao-22a}
Y.~Chu, J.~Yao, C.~Zhou, and H.~Yang, ``Graph neural networks in modern
  recommender systems,'' \emph{Graph Neural Networks: Foundations, Frontiers,
  and Applications}, pp. 423--445, 2022.

\bibitem{wan_yuy-22a}
Y.~Wang, Y.~Zhao, Y.~Zhang, and T.~Derr, ``Collaboration-aware graph
  convolutional networks for recommendation systems,'' \emph{arXiv preprint
  arXiv:2207.06221}, 2022.

\bibitem{ye_kum_sin-22a}
Z.~Ye, Y.~J. Kumar, G.~O. Sing, F.~Song, and J.~Wang, ``A comprehensive survey
  of graph neural networks for knowledge graphs,'' \emph{IEEE Access}, vol.~10,
  pp. 75\,729--75\,741, 2022.

\bibitem{yas_ren_bos-21a}
M.~Yasunaga, H.~Ren, A.~Bosselut, P.~Liang, and J.~Leskovec, ``Qa-gnn:
  Reasoning with language models and knowledge graphs for question answering,''
  \emph{arXiv preprint arXiv:2104.06378}, 2021.

\bibitem{zha_yao-22a}
Y.~Zhang and Q.~Yao, ``Knowledge graph reasoning with relational digraph,'' in
  \emph{Proceedings of the ACM Web Conference 2022}, 2022, pp. 912--924.

\bibitem{xia_wan_dai-22a}
S.~Xiao, S.~Wang, Y.~Dai, and W.~Guo, ``Graph neural networks in node
  classification: survey and evaluation,'' \emph{Machine Vision and
  Applications}, vol.~33, pp. 1--19, 2022.

\bibitem{oon_suz-19a}
K.~Oono and T.~Suzuki, ``Graph neural networks exponentially lose expressive
  power for node classification,'' \emph{arXiv preprint arXiv:1905.10947},
  2019.

\bibitem{kip_wel-2016a}
T.~N. Kipf and M.~Welling, ``Semi-supervised classification with graph
  convolutional networks,'' \emph{arXiv preprint arXiv:1609.02907}, 2016.

\bibitem{vel_pet_gui-2017a}
P.~Veli{\v{c}}kovi{\'c}, G.~Cucurull, A.~Casanova, A.~Romero, P.~Lio, and
  Y.~Bengio, ``Graph attention networks,'' \emph{arXiv preprint
  arXiv:1710.10903}, 2017.

\bibitem{ham_wil_jur-2017a}
W.~Hamilton, Z.~Ying, and J.~Leskovec, ``Inductive representation learning on
  large graphs,'' \emph{Advances in neural information processing systems},
  vol.~30, 2017.

\bibitem{wij_wan-19a}
A.~Wijesinghe and Q.~Wang, ``Dfnets: Spectral cnns for graphs with
  feedback-looped filters,'' in \emph{Advances in Neural Information Processing
  Systems (NeurIPS)}, 2019.

\bibitem{yan_zha_zho-21a}
S.~Yang, Z.~Zhang, J.~Zhou, Y.~Wang, W.~Sun, X.~Zhong, Y.~Fang, Q.~Yu, and
  Y.~Qi, ``Financial risk analysis for smes with graph-based supply chain
  mining,'' in \emph{Proceedings of the Twenty-Ninth International Conference
  on International Joint Conferences on Artificial Intelligence}, 2021, pp.
  4661--4667.

\bibitem{hu_fey_zit-20a}
W.~Hu, M.~Fey, M.~Zitnik, Y.~Dong, H.~Ren, B.~Liu, M.~Catasta, and J.~Leskovec,
  ``Open graph benchmark: Datasets for machine learning on graphs,''
  \emph{Advances in neural information processing systems}, vol.~33, pp.
  22\,118--22\,133, 2020.

\bibitem{lia_yan_hu-20a}
M.~Liang, B.~Yang, R.~Hu, Y.~Chen, R.~Liao, S.~Feng, and R.~Urtasun, ``Learning
  lane graph representations for motion forecasting,'' in \emph{Computer
  Vision--ECCV 2020: 16th European Conference, Glasgow, UK, August 23--28,
  2020, Proceedings, Part II 16}.\hskip 1em plus 0.5em minus 0.4em\relax
  Springer, 2020, pp. 541--556.

\bibitem{aga_lak_zit-21a}
C.~Agarwal, H.~Lakkaraju, and M.~Zitnik, ``Towards a unified framework for fair
  and stable graph representation learning,'' in \emph{Uncertainty in
  Artificial Intelligence}.\hskip 1em plus 0.5em minus 0.4em\relax PMLR, 2021,
  pp. 2114--2124.

\bibitem{pan_nik_vaz-21a}
G.~Panagopoulos, G.~Nikolentzos, and M.~Vazirgiannis, ``Transfer graph neural
  networks for pandemic forecasting,'' in \emph{Proceedings of the AAAI
  Conference on Artificial Intelligence}, vol.~35, no.~6, 2021, pp. 4838--4845.

\bibitem{yu_lia-23a}
J.~Yu, J.~Liang, and R.~He, ``Mind the label shift of augmentation-based graph
  ood generalization,'' in \emph{Proceedings of the IEEE/CVF Conference on
  Computer Vision and Pattern Recognition}, 2023, pp. 11\,620--11\,630.

\bibitem{sui_wan_jia-22a}
Y.~Sui, X.~Wang, J.~Wu, A.~Zhang, and X.~He, ``Adversarial causal augmentation
  for graph covariate shift,'' \emph{arXiv preprint arXiv:2211.02843}, 2022.

\bibitem{fen_wen_jie-20a}
W.~Feng, J.~Zhang, Y.~Dong, Y.~Han, H.~Luan, Q.~Xu, Q.~Yang, E.~Kharlamov, and
  J.~Tang, ``Graph random neural networks for semi-supervised learning on
  graphs,'' \emph{Advances in neural information processing systems}, vol.~33,
  pp. 22\,092--22\,103, 2020.

\bibitem{li_hao_ziw-22a}
H.~Li, Z.~Zhang, X.~Wang, and W.~Zhu, ``Disentangled graph contrastive learning
  with independence promotion,'' \emph{IEEE Transactions on Knowledge and Data
  Engineering}, 2022.

\bibitem{fan_wan_mo-2022a}
S.~Fan, X.~Wang, Y.~Mo, C.~Shi, and J.~Tang, ``Debiasing graph neural networks
  via learning disentangled causal substructure,'' \emph{arXiv preprint
  arXiv:2209.14107}, 2022.

\bibitem{li_hao_ziw-21a}
H.~Li, X.~Wang, Z.~Zhang, Z.~Yuan, H.~Li, and W.~Zhu, ``Disentangled
  contrastive learning on graphs,'' \emph{Advances in Neural Information
  Processing Systems}, vol.~34, pp. 21\,872--21\,884, 2021.

\bibitem{zho_kut_rib-22a}
Y.~Zhou, G.~Kutyniok, and B.~Ribeiro, ``Ood link prediction generalization
  capabilities of message-passing gnns in larger test graphs,'' \emph{arXiv
  preprint arXiv:2205.15117}, 2022.

\bibitem{li_hao_ziw-22b}
H.~Li, Z.~Zhang, X.~Wang, and W.~Zhu, ``Learning invariant graph
  representations for out-of-distribution generalization,'' in \emph{Advances
  in Neural Information Processing Systems}, 2022.

\bibitem{buf_dav_pie-22a}
D.~Buffelli, P.~Li{\`o}, and F.~Vandin, ``Sizeshiftreg: a regularization method
  for improving size-generalization in graph neural networks,'' \emph{arXiv
  preprint arXiv:2207.07888}, 2022.

\bibitem{wu_boj_ale-22a}
Y.~Wu, A.~Bojchevski, and H.~Huang, ``Adversarial weight perturbation improves
  generalization in graph neural network,'' \emph{arXiv preprint
  arXiv:2212.04983}, 2022.

\bibitem{sad_ma_li-21a}
A.~Sadeghi, M.~Ma, B.~Li, and G.~B. Giannakis, ``Distributionally robust
  semi-supervised learning over graphs,'' \emph{arXiv preprint
  arXiv:2110.10582}, 2021.

\bibitem{fen_he_tan-19a}
F.~Feng, X.~He, J.~Tang, and T.-S. Chua, ``Graph adversarial training:
  Dynamically regularizing based on graph structure,'' \emph{IEEE Transactions
  on Knowledge and Data Engineering}, vol.~33, no.~6, pp. 2493--2504, 2019.

\bibitem{wan_li_jin-22a}
Y.~Wang, C.~Li, W.~Jin, R.~Li, J.~Zhao, J.~Tang, and X.~Xie, ``Test-time
  training for graph neural networks,'' \emph{arXiv preprint arXiv:2210.08813},
  2022.

\bibitem{liu_hu_wan-22a}
H.~Liu, B.~Hu, X.~Wang, C.~Shi, Z.~Zhang, and J.~Zhou, ``Confidence may cheat:
  Self-training on graph neural networks under distribution shift,'' in
  \emph{Proceedings of the ACM Web Conference 2022}, 2022, pp. 1248--1258.

\bibitem{liu_jin_pan-22a}
Y.~Liu, M.~Jin, S.~Pan, C.~Zhou, Y.~Zheng, F.~Xia, and S.~Y. Philip, ``Graph
  self-supervised learning: A survey,'' \emph{IEEE Transactions on Knowledge
  and Data Engineering}, vol.~35, no.~6, pp. 5879--5900, 2022.

\bibitem{liu_che_yan-18a}
Z.~Liu, C.~Chen, X.~Yang, J.~Zhou, X.~Li, and L.~Song, ``Heterogeneous graph
  neural networks for malicious account detection,'' in \emph{Proceedings of
  the 27th ACM international conference on information and knowledge
  management}, 2018, pp. 2077--2085.

\bibitem{wan_lin_cui-19a}
D.~Wang, J.~Lin, P.~Cui, Q.~Jia, Z.~Wang, Y.~Fang, Q.~Yu, J.~Zhou, S.~Yang, and
  Y.~Qi, ``A semi-supervised graph attentive network for financial fraud
  detection,'' in \emph{2019 IEEE International Conference on Data Mining
  (ICDM)}.\hskip 1em plus 0.5em minus 0.4em\relax IEEE, 2019, pp. 598--607.

\bibitem{zhu-21a}
Q.~Zhu, N.~Ponomareva, J.~Han, and B.~Perozzi, ``Shift-robust gnns: Overcoming
  the limitations of localized graph training data,'' \emph{Advances in Neural
  Information Processing Systems}, vol.~34, 2021.

\bibitem{gas_etal-19a}
J.~Gasteiger, A.~Bojchevski, and S.~G{\"u}nnemann, ``Predict then propagate:
  Graph neural networks meet personalized pagerank,'' in \emph{International
  Conference on Learning Representations (ICLR)}, 2019.

\bibitem{liu_gao_ji-20a}
M.~Liu, H.~Gao, and S.~Ji, ``Towards deeper graph neural networks,'' in
  \emph{Proceedings of the 26th ACM SIGKDD international conference on
  knowledge discovery \& data mining}, 2020, pp. 338--348.

\end{thebibliography}
\end{document}